\documentclass[12pt]{article}
\usepackage{fontspec}
\usepackage[protrusion=true]{microtype}
\usepackage{geometry}
\usepackage{graphicx}
\usepackage{amsmath,amsthm,mathtools}
\usepackage{booktabs,tabularx,array}
\usepackage[dvipsnames]{xcolor}
\usepackage[numbers,sort&compress]{natbib}
\usepackage{xurl}
\usepackage[hidelinks]{hyperref}
\hypersetup{pdftitle={Conditional Weight of Evidence for Gradient-Boosted Credit Models},
 pdfauthor={Agus Sudjianto and Denis Burakov}}
\usepackage{setspace}
\usepackage[font=small,labelfont=bf,skip=6pt]{caption}
\usepackage{titlesec}
\titlespacing*{\section}{0pt}{1.2em}{0.6em}
\titlespacing*{\subsection}{0pt}{1em}{0.5em}
\titlespacing*{\subsubsection}{0pt}{0.9em}{0.4em}
\usepackage{enumitem}
\usepackage{placeins}

\usepackage{unicode-math}
\setlist{leftmargin=2em, itemsep=0.25em, topsep=0.5em, partopsep=0pt}
\newtheorem{proposition}{Proposition}
\newtheorem{definition}{Definition}
\newcommand{\Prb}{\mathbb P}
\newcommand{\E}{\mathbb E}
\newcommand{\ind}{\mathbf 1}
\newcommand{\logit}{\operatorname{logit}}
\newcommand{\sig}{\sigma}
\newcolumntype{Y}{>{\raggedright\arraybackslash}X}
\newcolumntype{R}{>{\raggedleft\arraybackslash}X}

\title{Conditional Weight of Evidence for Gradient-Boosted Credit Models:\\[0.3em]
\large Estimation, Additivity and Its Relation to Adverse Action Reasons}
\author{Agus Sudjianto \and Denis Burakov}
\date{October 2026}

\begin{document}
\onehalfspacing
\maketitle

\begin{abstract}
\setstretch{1}
\noindent
Traditional weight of evidence assigns a credit characteristic a value based on its association with default across a population. Credit analysis also requires an assessment of that characteristic in the context of an individual applicant's record. We study conditional weight of evidence for this purpose. It measures the change in default log odds from observing a characteristic after the remaining credit information is known. Its sign identifies whether the characteristic adds adverse or favorable information for that applicant, and its magnitude measures the strength of that information. The same characteristic value can therefore receive different assessments for applicants with different payment histories or other credit attributes. We develop estimators using gradient-boosted trees and evaluate their accuracy in a bureau-style simulation. Regressions of the full model's predicted probabilities recover conditional evidence more accurately than separate models fitted to default outcomes, provided the full model is accurate. They also recover a deliberately constructed file-age effect whose direction varies with utilization and is often missed by marginal weight of evidence. Local class-frequency estimates improve on marginal weight of evidence but remain inaccurate for some weak effects. An application to FICO HELOC data illustrates estimation when the population evidence is unknown. Conditional weight of evidence provides an applicant-specific interpretation of the information in a credit characteristic; its population summary, a conditional information value, splits the joint information in the record across blocks when averaged over orderings. Neither is an adverse action reason, which explains the fitted score difference between the applicant and chosen passing records.

\end{abstract}

\section{Introduction}

When an application is declined on the basis of a credit score, the bank needs to explain which characteristics of the credit record account for the unfavorable score. A familiar adverse characteristic, such as high revolving utilization, is not sufficient by itself to explain the decision. Its contribution depends on how the model uses that characteristic and on the comparison used to assess the applicant's score. For a credit analyst, the explanation should connect the scored disadvantage to identifiable features of the record, such as recent payment problems or heavy use of available credit.

A conventional logistic scorecard makes this connection through characteristic points. Categories of utilization, delinquency and file age receive weight-of-evidence (WoE) values, which are multiplied by fitted regression coefficients to obtain their contributions to the score \citep{siddiqi,good}. WoE summarizes how common a category is among defaulters relative to non-defaulters. Its value is calculated for the category across the development population: utilization of 85 percent receives the same WoE value in a fifteen-year file with no delinquency as in a two-year file with recent past-due accounts. The regression coefficients account for the other characteristics in the scorecard, but they do not make the underlying WoE value specific to the applicant's remaining record.

Gradient-boosted decision trees (GBDTs) can allow the relationship between utilization and default risk to vary with payment history, file age and other characteristics. Interpreting such a model requires distinguishing a characteristic's contribution to the fitted score from the additional information it supplies about default. Delinquency, inquiries and utilization can convey overlapping information about financial stress. Once the payment history and inquiries are known, utilization may add relatively little information even when the model assigns it a substantial contribution to the applicant's score disadvantage. A marginal WoE table cannot establish how much information remains in that particular context.

Conditional WoE addresses this question by comparing default odds with and without knowledge of the characteristic, given the rest of the credit record. The comparison without revolving information averages over the revolving profiles found among applicants with the same remaining characteristics. It does not assume a zero balance or an unused credit line. The averaging is done on the probability scale, before conversion to log odds; averaging log odds instead, as Shapley values of a model's log-odds output do, gives a different quantity (Proposition~\ref{prop:modelgame}). A positive conditional WoE means that learning the applicant's revolving profile raises default odds relative to that comparison; a negative value means that it lowers them. We assess balance, limit and utilization together because they describe the same exposure and are linked by an accounting identity. This grouping avoids treating utilization as additional information when balance and limit already determine it.

Explaining the score shortfall involves a further choice: the applicants against whom the declined record is compared. Passing applicants provide observed credit profiles with more favorable scores, and an attribution allocates the score difference across characteristics. Regulation B and its Official Interpretations describe comparisons with applicants at or slightly above the passing score, as well as comparisons with all applicants, among the methods for selecting reasons. The reasons must relate to the factors actually considered or scored; the creditor must provide a statement of specific reasons or disclose the applicant's right to obtain one \citep{regb,interp}. The passing comparison and the conditional-risk comparison serve different purposes. One accounts for a difference between scored records; the other measures the default information added by a characteristic after the remaining record is known.

The distinction has a concrete consequence in our simulation. An applicant with a recent serious delinquency and revolving utilization of 53 percent has delinquency as the largest contribution to the score gap against a broad sample of near-passing applicants. Among passing applicants with more similar delinquency histories, mean utilization is 27 percent and revolving credit becomes the largest contribution. The applicant's record and the fitted model have not changed. The change in the leading factor follows from the credit profiles used for comparison. Section~\ref{sec:case} examines this case, including the favorable characteristics that offset part of the applicant's disadvantage.

We estimate conditional WoE using separate models fitted with and without a characteristic, regressions of the full model's predicted probabilities and class frequencies among similar records. A bureau-style simulation allows their errors to be measured against numerically computed population evidence. We also examine how overlapping information affects the allocation of total evidence and how restricting performance data to booked accounts changes the comparison. The FICO HELOC application examines the estimated evidence in actual credit records and its relation to attribution against passing applicants.

Applicant-level evidence also needs a population summary. The mean of $W_j$ over defaulters minus its mean over non-defaulters gives a conditional information value, the information a block adds once the rest of the record is known: the familiar information value with conditional evidence in place of marginal WoE. Its order-averaged version splits the joint information value of the record exactly across blocks.

In the simulation, regressions of the full model's probabilities estimate applicant-level conditional WoE more accurately than separate models fitted to default outcomes. The estimates recover differences in revolving and delinquency evidence across applicants, while weak effects remain harder to distinguish from estimation error. The file-age example shows how a characteristic can supply favorable or adverse evidence depending on the rest of the record. The passing-reference comparisons then distinguish this local risk information from the characteristic's contribution to a particular score gap.

\section{Related work}\label{sec:related}

Good's log likelihood ratio measures evidence for one outcome relative to another \citep{good}. Scorecard WoE applies this idea to a characteristic class before fitting a logistic regression \citep{siddiqi}. The regression estimates joint score coefficients, but the class WoE itself remains a marginal population summary. The conditional definition below asks what the observed class or block adds after specified other characteristics are known.

Shapley attribution allocates a model output or contrast by averaging a characteristic's contribution over possible orders of entry \citep{lundberg,treeshap}. The result depends on what replaces the absent characteristics \citep{baseline,covert}. An observational construction averages over their conditional distribution; an interventional construction draws them independently of the observed characteristics. Both use a model and a distribution, but they answer different comparison questions \citep{janzing,truedata,aas}. Baseline Shapley fixes absent characteristics to a reference record \citep{baseline}. Nair et al.\ use passing references to explain adverse credit decisions \citep{nair}; companion manuscripts examine how leaf proximity and kernel similarity change that reference population \citep{aarcgbdt,aarcneural}. We retain this fitted-score comparison while studying conditional default evidence separately.

The direct estimator is a pointwise form of leave-one-covariate-out (LOCO) importance \citep{loco}, which compares a predictor fitted with and without a covariate. Removal by retraining is one of the removal mechanisms catalogued by \citet{covert}. The leaf-local estimator builds on the proximity kernel of random forests \citep{breiman}, the fraction of trees in which two records share a terminal node, and on the use of forest-derived weights for local estimation in generalized random forests \citep{grf}. Unlike those uses, the target here is a ratio of class-conditional distributions of a held-out characteristic, and the kernel is constructed from an ensemble that excludes that characteristic.

\section{Setup and notation}\label{sec:setup}

Let $Y=1$ denote default within the performance window and $Y=0$ its absence. The applicant's credit record $X=(X_1,\ldots,X_J)$ is divided into $J$ characteristic blocks. A block may contain several fields that must be interpreted together. Revolving limit, balance and utilization form one block because utilization is the ratio of the other two, and replacing one without the others would produce a record that no applicant could have. For a set $S\subseteq[J]=\{1,\ldots,J\}$, $x_S$ denotes the fields of the blocks in $S$, and $x_{-j}$ denotes all blocks except $j$.

The production model is a GBDT whose margin is recalibrated to estimated default log odds $f(x)$. An application is declined when $f(x)>c$. The adverse action comparison uses passing references $r$ with $c-\delta<f(r)\le c$, observed applicants whose estimated risk is at or just below the cutoff. The applicant's excess estimated default log odds relative to reference $r$ is
\begin{equation}\label{eq:gap}
D_f(x,r)=f(x)-f(r)>0 .
\end{equation}
An attribution of the fitted decision allocates $D_f$ across blocks. A block with a positive allocation raises the applicant's estimated default odds relative to the reference.

Evidence is defined from the population distribution instead of the fitted model. Throughout, the conditional probabilities $\Prb(Y=1\mid x_S)$ are assumed to lie strictly between zero and one, and the conditional densities of the blocks are assumed positive for both outcome classes on the records considered, so that every log ratio below is finite. With $\pi=\Prb(Y=1)$, the weight of evidence of the whole record is
\begin{equation}\label{eq:joint}
W(x)=\log\frac{p(x\mid Y=1)}{p(x\mid Y=0)}=\logit\Prb(Y=1\mid x)-\logit\pi ,
\end{equation}
which follows from Bayes' rule. Throughout, $\log$ denotes the natural logarithm, so evidence and log odds are in natural units. If $f$ equals the true default log odds, then $f(x)-f(r)=W(x)-W(r)$. Otherwise the two differences need not agree.

\section{Conditional weight of evidence}\label{sec:evidence}

\subsection{Definition}

Suppose, as a numerical illustration, that applicants with the same remaining credit characteristics have a default probability of 10 percent. Including the applicant's revolving information raises that probability to 20 percent. The default odds rise from $0.10/0.90$ to $0.20/0.80$, a factor of 2.25. The revolving block supplies $\log(2.25)=0.81$ units of conditional evidence. This is a comparison of risk assessments before and after the information is known. It does not specify a passing applicant or describe the effect of an intervention on balance.

For block $j$, the corresponding population quantity is
\begin{equation}\label{eq:cwoe}
W_j(x)=\log\frac{p(x_j\mid Y=1,x_{-j})}{p(x_j\mid Y=0,x_{-j})}=\logit\Prb(Y=1\mid x)-\logit\Prb(Y=1\mid x_{-j}).
\end{equation}
The second equality follows from Bayes' rule. The reduced probability averages over the values of block $j$ that occur among records with the same $x_{-j}$; leaving the block unobserved does not mean setting its fields to zero or to a missing-value code. A positive $W_j$ means the observed block raises default odds relative to that conditional average, and a negative value means it lowers them. The prior $\pi$ cancels from the likelihood ratio. Values near zero occur when the block is unremarkable given the rest of the record or when the other fields already determine it. If utilization were separated from balance and limit, for example, it would add no information because those fields determine it.

The definition therefore depends on the grouping of fields. Within-block identities are preserved by treating balance, limit and utilization together. For local estimation, the comparison records must be selected using the remaining blocks: a neighborhood built from the full model can also select on the block whose evidence is being estimated. That selection changes the comparison distribution.

\subsection{Additivity}

Two duplicate indicators of the same delinquency explain why the leave-one-block-out values need not add. Removing either indicator leaves the information in the other, so each has zero conditional evidence even though the delinquency is informative. By contrast, a pair of characteristics can be informative only when observed together. Removing either then removes the joint information. Regression-weighted scorecard points add to a score by its specification; these conditional evidence values have no such constraint.

\begin{proposition}[Non-additivity in both directions]\label{prop:nonadd}
The sum $\sum_jW_j(x)$ can be smaller or larger in absolute value than $W(x)$.
\end{proposition}
\begin{proof}
Redundancy. Let $X_1=X_2=Z$ for a binary $Z$ with $W(z)\neq0$ for some $z$. Each block is determined by the other, so $p(x_1\mid y,x_2)=1$ for both classes on the support and $W_1=W_2=0$, while $W(x)=W(z)\neq0$.
Synergy. Let $X_1,X_2$ be independent fair bits and let $\Prb(Y=1\mid x)=\sig(a+b\,\ind\{x_1\neq x_2\})$ with $b\neq0$. Then $\Prb(Y=1\mid x_1)$ and $\Prb(Y=1\mid x_2)$ both equal $\pi$, so $W_1(x)=W_2(x)=W(x)$ and $\sum_jW_j(x)=2W(x)$.
\end{proof}

The two examples show that the discrepancy can take either sign. They do not supply a rule that classifies an applicant as redundant or synergistic from the sign of $\sum_jW_j(x)-W(x)$, because that sign also depends on the sign of $W(x)$ and on how dependence and interaction combine across several blocks. Section~\ref{sec:results} reports how often each sign occurs in the simulation.

An allocation of total evidence must also assign information that several blocks share. Imagine revealing the blocks one at a time. In the duplicate-indicator example, the first indicator supplies the delinquency evidence and the second adds nothing. Reversing the order reverses their allocations; averaging the two orders divides the evidence equally. For an ordering $\rho$ with $S_{k-1}=\{\rho(1),\ldots,\rho(k-1)\}$, the general chain rule, which is Good's additive property of weight of evidence \citep{good}, is
\begin{equation}\label{eq:chain}
W(x)=\sum_{k=1}^J\log\frac{p(x_{\rho(k)}\mid Y=1,x_{S_{k-1}})}{p(x_{\rho(k)}\mid Y=0,x_{S_{k-1}})} .
\end{equation}
The total does not depend on the ordering, but the term assigned to each block does. Averaging over orderings removes that dependence. Define the coalition evidence $V_x(S)=\logit\Prb(Y=1\mid x_S)-\logit\pi$, the evidence of the blocks in $S$ alone, with $V_x(\varnothing)=0$ and $V_x([J])=W(x)$. The Shapley value of block $j$ in this game,
\begin{equation}\label{eq:shapley_evidence}
E_j(x)=\sum_{S\subseteq[J]\setminus\{j\}}\frac{|S|!\,(J-|S|-1)!}{J!}\bigl[V_x(S\cup\{j\})-V_x(S)\bigr],
\end{equation}
is the average over orderings of the chain-rule term of block $j$, and $\sum_jE_j(x)=W(x)$ by the efficiency property. Each increment $V_x(S\cup\{j\})-V_x(S)$ is the conditional evidence of block $j$ given only the blocks in $S$, and $W_j(x)$ is the increment for $S=[J]\setminus\{j\}$. We call
\begin{equation}\label{eq:shared}
H_j(x)=E_j(x)-W_j(x)
\end{equation}
the coalition excess of block $j$: its evidence averaged over all orders of entry minus its evidence when it enters last. In the redundancy example, $E_1=E_2=W(x)/2$ and $W_1=W_2=0$, so $H_j=W(x)/2$, which is negative for a favorable record with $W(x)<0$. In the synergy example, $E_j=W(x)/2$ and $W_j=W(x)$, so $H_j=-W(x)/2$. The sign of $H_j$ therefore does not by itself identify redundancy or synergy, and it is used below only as a defined contrast between two allocations of the same evidence. The Shapley construction is standard; what is specific here is the game, whose value function is a population log likelihood ratio.

\subsection{Relation to model Shapley values}

A model can recover the same evidence allocation if its probabilities describe the population and absent characteristics are averaged over the appropriate conditional distribution. The averaging scale matters: the log odds of an average default probability need not equal the average log odds.

\begin{proposition}\label{prop:modelgame}
Let $f^\ast(x)=\logit\Prb(Y=1\mid x)$ and define the observational model game $v(S)=\logit\E[\sig(f^\ast(X))\mid X_S=x_S]$. Then $v(S)-v(\varnothing)=V_x(S)$ for every $S$, and the Shapley values of $v$ equal $E_j(x)$. The game $v'(S)=\E[f^\ast(X)\mid X_S=x_S]$, which averages log odds instead of probabilities, differs from $v$ in general, and interventional or baseline games, which replace $X_{-S}$ by values not drawn from its conditional distribution, also differ from $v$ in general.
\end{proposition}
\begin{proof}
By the tower property, $\E[\sig(f^\ast(X))\mid x_S]=\E[\Prb(Y=1\mid X)\mid x_S]=\Prb(Y=1\mid x_S)$, so $v(S)=\logit\Prb(Y=1\mid x_S)$ and $v(\varnothing)=\logit\pi$. Shapley values are unchanged by adding a constant to the game. For the second game, suppose $f^\ast(X)$ takes the values $0$ and $4$ with equal probability given $x_S$. Then $v'(S)=2$, while $v(S)=\logit\{(\sig(0)+\sig(4))/2\}=1.05$.
\end{proof}

Proposition~\ref{prop:modelgame} gives sufficient conditions under which evidence and attribution agree: the model equals the true posterior, absent blocks are integrated over their conditional distribution and the integration is done on the probability scale. It does not show that agreement is impossible otherwise. The attribution game used for adverse action in Section~\ref{sec:aarc} can be evaluated with the true posterior, as in the simulation below, but it sets the absent blocks to the values of a passing reference and averages on the log-odds scale. It therefore differs from the coalition evidence game in its reference distribution and its averaging scale.

\subsection{Conditional information value}\label{sec:civ}

Information value is the population summary of weight of evidence: the average WoE among defaulters minus the average among non-defaulters. For a characteristic grouped into categories $b$, with shares $p_{1b}$ and $p_{0b}$ among defaulters and non-defaulters, it is $\sum_b(p_{1b}-p_{0b})\,\mathrm{WoE}_b$. The two averages are $\sum_bp_{1b}\mathrm{WoE}_b=\mathrm{KL}(p_1\|p_0)$ and $\sum_bp_{0b}(-\mathrm{WoE}_b)=\mathrm{KL}(p_0\|p_1)$, and their sum is the Jeffreys divergence \citep{jeffreys,sbinfo}, written here without the factor $\tfrac12$. Since $W_j$ is the WoE of a block for one applicant, averaging it in the same way gives the conditional version.

\begin{definition}
The conditional information value of block $j$ is
\begin{equation}\label{eq:civ}
\mathrm{IV}_j=\E[W_j(X)\mid Y=1]-\E[W_j(X)\mid Y=0],
\end{equation}
the average conditional evidence among defaulters minus its average among non-defaulters. The coalition information value $\mathrm{IV}^E_j$ is defined in the same way with $E_j$ in place of $W_j$.
\end{definition}

It measures how much a block tells us about default once the rest of the record is known: by the same decomposition, it averages over the rest of the record the two divergences between the distributions of $X_j$ among defaulters and among non-defaulters. Because it is a difference of two averages of evidence, it is on the WoE scale, in natural log-odds units, and can be compared directly with the marginal information value of the same block.

\begin{proposition}\label{prop:civ}
(i) $\mathrm{IV}_j\ge0$, with equality if and only if $X_j$ and $Y$ are conditionally independent given $X_{-j}$. (ii) The coalition values add up to the information value of the whole record:
\begin{equation}\label{eq:civ_sum}
\sum_j\mathrm{IV}^E_j=\E[W(X)\mid Y=1]-\E[W(X)\mid Y=0].
\end{equation}
\end{proposition}
\begin{proof}
(i) For fixed $x_{-j}$, the average of $W_j$ given $Y=1$ is $\mathrm{KL}\{p(\cdot\mid Y=1,x_{-j})\,\|\,p(\cdot\mid Y=0,x_{-j})\}$, and the average of $-W_j$ given $Y=0$ is the same divergence with the classes exchanged. Both are nonnegative and zero only when the two distributions coincide. (ii) The coalition values add up to the total evidence, $\sum_jE_j=W$, and the class mean is linear.
\end{proof}

Along any ordering, the class averages of the chain-rule terms in Eq.~\eqref{eq:chain} are nonnegative for the same reason and add up to the total. $\mathrm{IV}_j$ is the term of block $j$ when it enters last, and $\mathrm{IV}^E_j$ is the average of its terms over orderings.

Since $W=f^\ast-\logit\pi$, the total in Eq.~\eqref{eq:civ_sum} is also $\E[f^\ast(X)\mid Y=1]-\E[f^\ast(X)\mid Y=0]$, the gap between the average log odds of defaulters and non-defaulters, and for a calibrated model it can be estimated by $\E[f(X)\mid Y=1]-\E[f(X)\mid Y=0]$ on held-out records. Marginal values count shared information more than once, and neither they nor the leave-one-block-out values add up to the total in general: in the duplicate-indicator example of Proposition~\ref{prop:nonadd}, $\mathrm{IV}_1=\mathrm{IV}_2=0$ although the joint value is positive, and in the interaction example $\mathrm{IV}_1+\mathrm{IV}_2$ is twice the joint value.

In practice $\mathrm{IV}_j$ is the difference of the class means of $\widehat W_j$ on held-out records. When the evidence is constant within cells, this is the information-value sum over cells with conditional evidence in place of WoE, with the evidence and the cell shares taken from different samples, since computing both on the same records overstates it. The same definition applies to the class-level evidence $W^B_j$ and to the booked population.

\subsection{Booked and through-the-door populations}

A lender usually observes performance only for booked accounts. Even if the default probability for a fully specified record is unchanged by booking, the distribution of an omitted characteristic can change. A policy that screens out high utilization, for example, can leave lower utilization among booked applicants with the same other characteristics. This changes the reduced-risk comparison used in conditional evidence.

Suppose booking has probability $s(x)$ and is independent of the outcome given the observed record. Then $\Prb(Y=1\mid x,\text{booked})=\Prb(Y=1\mid x)$ wherever $s(x)>0$. This preserves a population posterior on that support; a fitted model still faces finite samples, restricted support and possible misspecification.

\begin{proposition}\label{prop:selection}
Under selection on the record, at a record with $s(x)>0$ and $\E[s(X)\mid x_{-j}]>0$, the conditional evidence of block $j$ in the booked population is
\begin{equation}\label{eq:booked}
W^{b}_j(x)=\logit\Prb(Y=1\mid x)-\logit\frac{\E[\Prb(Y=1\mid X)\,s(X)\mid x_{-j}]}{\E[s(X)\mid x_{-j}]},
\end{equation}
where the expectations are over $X_j$ given $x_{-j}$ in the through-the-door population. It equals $W_j(x)$ for all $x$ if $s$ does not depend on $x_j$, and more generally whenever $\Prb(Y=1\mid X)$ and $s(X)$ are uncorrelated given $x_{-j}$.
\end{proposition}
\begin{proof}
The booked density of $X_j$ given $x_{-j}$ is proportional to $p(x_j\mid x_{-j})s(x)$. Integrating $\Prb(Y=1\mid X)$ against it gives the second term. The ratio equals $\E[\Prb(Y=1\mid X)\mid x_{-j}]$ when the conditional covariance of $\Prb(Y=1\mid X)$ and $s(X)$ is zero.
\end{proof}

If, within the stratum defined by $x_{-j}$, the booking probability decreases as risk increases, then $\Prb(Y=1\mid X)$ and $s(X)$ are negatively correlated given $x_{-j}$, the booked conditional default rate given $x_{-j}$ is lower and $W^b_j$ exceeds $W_j$. A lower overall default rate among booked accounts does not by itself establish this conditional relationship. The difference is confined to blocks on which $s$ depends, and it does not vanish for applicants inside the booked support, because the integral runs over values of $X_j$ that the policy would have declined.

\section{Estimators}\label{sec:estimation}

\subsection{Direct estimator}

To assess what revolving credit adds, the direct estimator compares default risk fitted from the full record with risk fitted from the remaining characteristics. The reduced model describes applicants whose delinquency, file age and other blocks are known but whose revolving information is not supplied. Both models are fitted to the same outcome in the same population and recalibrated by Platt scaling on a separate calibration sample \citep{platt}. Their log-odds difference is
\begin{equation}\label{eq:direct}
\widehat W_j^{\rm dir}(x)=\logit\hat p_F(x)-\logit\hat p_{-j}(x_{-j}).
\end{equation}
The reduced model is refitted without the block. Supplying missing values to the full model would instead use branches learned for missing data, which need not represent an average over the omitted characteristic. The two fits have separate estimation errors and can differ even when the block adds no population evidence. Probability evaluations are clipped to $[10^{-4},1-10^{-4}]$ where converted numerically to log odds; calibrated margins used directly retain their log-odds values. The library comparison also excludes any derived categorical combination containing a removed field.

\subsection{Projection estimator}

The reduced risk can also be estimated by averaging the full model's probabilities among records with the same remaining characteristics. With the true posterior, this follows from the tower property:
\begin{equation}\label{eq:tower}
\Prb(Y=1\mid x_{-j})=\E\bigl[\Prb(Y=1\mid X)\mid X_{-j}=x_{-j}\bigr],
\end{equation}
The projection estimator uses the full model's probability as the regression target. A GBDT with the cross-entropy loss and soft targets $\hat p_F(X_i)\in(0,1)$ is fitted on the block-excluded fields of a sample not used to fit $\hat p_F$, giving $\tilde p_{-j}(x_{-j})$, and
\begin{equation}\label{eq:proj}
\widehat W_j^{\rm proj}(x)=\logit\hat p_F(x)-\logit\tilde p_{-j}(x_{-j}).
\end{equation}
The minimizer of expected cross-entropy with a soft target is the conditional mean of the target, so $\tilde p_{-j}$ estimates $\E[\hat p_F(X)\mid x_{-j}]$, which equals Eq.~\eqref{eq:tower} if $\E[\hat p_F(X)-\Prb(Y=1\mid X)\mid x_{-j}]=0$, in particular if $\hat p_F$ equals the true posterior. For the true posterior $p(X)=\Prb(Y=1\mid X)$,
\[
\operatorname{Var}(Y\mid x_{-j})=\E\bigl[p(X)(1-p(X))\mid x_{-j}\bigr]+\operatorname{Var}\bigl(p(X)\mid x_{-j}\bigr),
\]
and a soft target equal to $p(X)$ removes the first term. How much this reduces variance depends on the size of the first term relative to the second; the first term is small when the outcome is nearly determined by the record, even at low prevalence. With the estimated target $\hat p_F$, its estimation error enters the regression as well. Both terms of Eq.~\eqref{eq:proj} inherit the errors of $\hat p_F$, and the estimate is the conditional evidence implied by the full model and the population distribution of credit records.

\subsection{Error of a log-odds difference}

An accurate difference need not imply equally accurate probabilities in both terms. Let $e_F=\logit\hat p_F-\logit p$ be the full-model error and $e_{-j}$ the error of the reduced log odds, using either estimator. At the same record,
\begin{equation}\label{eq:error_difference}
\widehat W_j-W_j=e_F-e_{-j},\qquad
\E[(\widehat W_j-W_j)^2]=\E[e_F^2]+\E[e_{-j}^2]-2\E[e_Fe_{-j}].
\end{equation}
The expectations here are over the evaluation records. A smaller reduced-model error improves the difference, but errors in the same direction can also cancel. Projection learns from the full model and can share its errors. The experiments report both reduced-posterior error and the cross-product term to distinguish these effects.

\subsection{Leaf-local estimator}

A frequency-based estimate asks how often the applicant's class of a characteristic occurs among defaulters and non-defaulters with similar remaining records. The challenge is obtaining a sufficiently local comparison with enough observations in both outcome classes. We use a block-excluded tree ensemble $g_{-j}$ to weight similarity in the remaining blocks.

This estimator first groups block values into classes. For a multi-field block, the classes are quantile intervals of a logistic index fitted from transformed block fields on the development sample. If $b(x_j)$ is the applicant's class, the target is
\begin{equation}\label{eq:binwoe}
W_j^{B}(x)=\log\frac{\Prb(X_j\in b(x_j)\mid Y=1,x_{-j})}{\Prb(X_j\in b(x_j)\mid Y=0,x_{-j})},
\end{equation}
This class-level quantity averages over values within the class. It agrees throughout a class with Eq.~\eqref{eq:cwoe} when the conditional density ratio is constant there; otherwise coarsening changes the target.

Let $L_t(z)$ be the leaf reached by $z=x_{-j}$ in tree $t$ of $g_{-j}$, and let
\begin{equation}\label{eq:kernel}
K_{-j}(x,x')=\frac1T\sum_{t=1}^T\ind\{L_t(x_{-j})=L_t(x'_{-j})\}
\end{equation}
be the fraction of trees in which two records share a leaf \citep{breiman}. The kernel depends on $x$ only through $x_{-j}$. For an estimation sample disjoint from the samples used to fit $g_{-j}$ and choose the classes, define kernel-weighted counts
\begin{equation}\label{eq:counts}
n_{yb}(x)=\sum_iK_{-j}(x,x_i)^{\kappa}\,\ind\{Y_i=y,\,b(x_{ij})=b\},\qquad n_y(x)=\sum_bn_{yb}(x).
\end{equation}
With $\kappa=1$, $n_{yb}(x)$ is the number of estimation records of class $(y,b)$ sharing a leaf with the applicant, averaged over trees. Each root-to-leaf path has at most as many splits as the tree depth, so within a single leaf the records still differ in the fields not split on along that path. Averaging over trees does not remove that variation, while raising the kernel to a power $\kappa>1$ concentrates weight on records that share leaves with the applicant in many trees. The counts are on the scale of record counts, which fixes the scale of the smoothing constant. The estimate shrinks toward the class-specific global class frequencies $q^{(y)}_b=\Prb(b(X_j)=b\mid Y=y)$:
\begin{equation}\label{eq:leafwoe}
\begin{gathered}
\widehat p(b\mid y,x)=\frac{n_{yb}(x)+\lambda q^{(y)}_b}{n_y(x)+\lambda}=\frac{n_y(x)}{n_y(x)+\lambda}\,\frac{n_{yb}(x)}{n_y(x)}+\frac{\lambda}{n_y(x)+\lambda}\,q^{(y)}_b,\\
\widehat W_j^{\rm leaf}(x)=\log\frac{\widehat p(b(x_j)\mid1,x)}{\widehat p(b(x_j)\mid0,x)} .
\end{gathered}
\end{equation}
As $\lambda\to\infty$, the estimate tends to the global (marginal) WoE of the class, the scorecard quantity. A common smoothing distribution for both outcome classes would instead shrink the estimate toward zero. With $\lambda=0$, the estimate is a ratio of within-class frequencies and does not depend on the local default rate. With $\lambda>0$, the class totals determine how far each class-conditional frequency is shrunk, and the class with fewer local records, usually the defaulters, is shrunk more.

The exponent and smoothing constant balance locality against sampling noise. We select $(\kappa,\lambda)$ by the class-balanced held-out log likelihood $\tfrac12\sum_y\overline{\log\widehat p(b_i\mid y,x_i)}$ on the calibration sample, where the bar averages over records with outcome $y$. Equal weighting keeps the larger non-default class from dominating the criterion. No oracle evidence is used in this choice. The class-specific effective sample size, $\mathrm{ESS}_y(x)=(\sum_{i:Y_i=y}K_i^\kappa)^2/\sum_{i:Y_i=y}K_i^{2\kappa}$, measures how concentrated the weights are and is reported alongside the weighted counts.

Two geometries are compared. The first uses the block-excluded default model $g_{-j}$ from the direct estimator, whose leaves separate records by estimated risk given $x_{-j}$. The estimand, however, requires neighborhoods in which the class-conditional distribution of $X_j$ is approximately constant, and two contexts with similar risk can have different distributions of the excluded block. The second geometry is a multiclass GBDT fitted on the development sample to predict the class $b(X_j)$ from $x_{-j}$, whose leaves separate records by the distribution of the excluded block. Neither geometry is designed for the ratio in Eq.~\eqref{eq:binwoe} itself, and the comparison is empirical.

The leaves supply neighborhoods, not evidence values. Their fitted outputs depend on loss gradients and Hessians; a leaf output is not itself a class likelihood ratio.\footnote{From the base score $\logit\bar y$, a first-tree leaf weight $\sum_i(y_i-\bar y)/\{n_\ell\,\bar y(1-\bar y)\}$ is the first-order approximation of the WoE of its cell, $\logit\bar y_\ell-\logit\bar y$; later leaves approximate the evidence left after the earlier trees, and shrinkage and the $L_2$ penalty scale both down.} Evidence is obtained from the separate outcome-class frequencies in those neighborhoods.
\subsection{Coalition evidence}

Allocating evidence over all entry orders requires a reduced probability for every subset of blocks. A projection regression on $x_S$ estimates $\E[\hat p_F(X)\mid x_S]$, the full-model-implied counterpart of $\Prb(Y=1\mid x_S)$. We set $\widehat V_x(S)=\logit\tilde p_S(x_S)-\logit\hat\pi$ for proper nonempty subsets and use $\hat p_F$ for the full set, with $\widehat V_x(\varnothing)=0$. Here $\hat\pi$ is the evidence-sample default rate. Replacing it by the mean full-model probability shifts every nonempty coalition value by a common constant and each Shapley allocation by that constant divided by $J$. In the simulation the change is at most 0.007 logit units per block. Six blocks require 62 reduced regressions; the leave-one-block-out projections are the six subsets of size five. More blocks would require an approximation to the order average, which is not studied here.

\section{Adverse action reasons}\label{sec:aarc}

\subsection{Baseline Shapley attribution}

A passing reference supplies an observed alternative for each characteristic group. To attribute the applicant's score disadvantage, the comparison replaces whole blocks of the reference with those of the applicant and evaluates the change in the fitted score. Averaging a block's increment over replacement orders gives its baseline Shapley value \citep{baseline,nair}. Formally, $v_{x,r}(S)=f(x_S,r_{-S})-f(r)$, and the allocations $\phi_j^f(x,r)$ sum to $D_f(x,r)$. Averaging over references with nonnegative weights $\omega_r$ summing to one gives
\begin{equation}\label{eq:attr}
C^f_j(x)=\sum_{r\in\mathcal R(x)}\omega_r\,\phi^f_j(x,r),\qquad \sum_jC^f_j(x)=\sum_r\omega_rD_f(x,r).
\end{equation}
The block with the largest $C^f_j$ is the leading candidate reason. Exact reconciliation establishes that the allocation accounts for the specified contrast. It does not establish that the reference policy is appropriate, nor that a multi-field block identifies the field to be named in a notice: a positive revolving allocation does not by itself distinguish high utilization from a high balance or a low limit. Blocks are replaced whole, so each hybrid preserves the identities within a block. Plausibility of combinations across blocks is not checked. With six blocks, each reference requires 64 model evaluations, and $C^f$ is computed exactly. The group-level Shapley value of a block does not in general equal the sum of the feature-level Shapley values of its fields, so feature-level TreeSHAP output cannot in general be aggregated to obtain it.

Both reference policies use observed applicants in the passing band. Each declined applicant is compared with 50 references from the band (30 in HELOC). The score-band policy samples them uniformly. The proximity policy selects those sharing the most leaves with the applicant in the full model \citep{aarcgbdt}. The first asks how the applicant differs from the broader near-passing group; the second emphasizes records similar in the characteristics used by the fitted trees. Neither choice follows from the Shapley allocation rule.

For fixed applicants and references, positive affine recalibration $f=a+b\,m$ scales every block attribution by $b$: $C_j^f=bC_j^m$ for $b>0$. It leaves the leading block and the relative margin unchanged because the intercept cancels and all increments receive the same multiplier. The passing band has fixed width in calibrated log odds, however, so recalibration can change reference membership. Conditional evidence also responds differently: the direct method uses separately recalibrated fits and projection averages probabilities before taking log odds.

\subsection{Evidence comparisons with passing references}

Attribution compares the applicant with passing references. Conditional evidence compares the applicant's block with the blocks found among records that share the applicant's other characteristics. A reference-based form of conditional evidence evaluates the evidence of each reference's block in the applicant's context:
\begin{equation}\label{eq:comparative}
C^W_j(x)=\sum_r\omega_r\bigl[\widehat W_j(x_j;x_{-j})-\widehat W_j(r_j;x_{-j})\bigr].
\end{equation}
This holds the applicant's other blocks fixed and varies one block, as one step of the attribution game does. For the direct and projection estimators both terms use the same reduced-risk prediction, which cancels in the difference.

\begin{proposition}\label{prop:collapse}
If $\widehat W_j(x_j;x_{-j})=\hat g(x_j,x_{-j})-\hat h(x_{-j})$ for some functions $\hat g$ and $\hat h$, then
\[
C^W_j(x)=\sum_r\omega_r\bigl[\hat g(x_j,x_{-j})-\hat g(r_j,x_{-j})\bigr],
\]
and $\hat h$ does not enter. For the direct and projection estimators, $\hat g=\logit\hat p_F$, so $C^W_j$ is the average change in the full model's log odds when block $j$ of the applicant is replaced by the reference's block. For the leaf-local estimator, $C^W_j$ is the average difference in log ratios of the smoothed class-conditional frequencies of the applicant's and the reference's classes, computed in the applicant's context, in which the normalizers $n_y(x)+\lambda$ cancel.
\end{proposition}
\begin{proof}
Both terms of Eq.~\eqref{eq:comparative} use the same context $x_{-j}$, so $\hat h(x_{-j})$ cancels. In the leaf case, both terms of Eq.~\eqref{eq:leafwoe} are evaluated at the same $x$, so $n_1(x)+\lambda$ and $n_0(x)+\lambda$ cancel from the difference.
\end{proof}

When $\hat p_F$ is the production model, the direct and projection forms of $C^W_j$ reduce to $v_{x,r}([J])-v_{x,r}([J]\setminus\{j\})$, the increment of block $j$ when it is replaced last in the attribution game. They describe the fitted model, not default evidence, and differ from $C^f_j$ only because the Shapley value averages this increment with the increments at other orders of entry. The leaf-local form retains class frequencies estimated from outcomes and does not reduce to a model swap, but it compares coarse classes rather than records. Conditional evidence is therefore reported for the applicant alone, as $W_j(x)$ and $E_j(x)$, alongside the attribution against passing references.

\subsection{Hypotheses}

The simulation permits direct assessment of evidence accuracy because its population quantities can be computed numerically. The real-data study examines estimator agreement, local sample support and the interpretation of evidence alongside score attribution. The empirical hypotheses are:
\begin{itemize}
\item[H1.] The direct and projection estimators recover the sign and magnitude of $W_j(x)$, and the leaf-local estimator recovers $W^B_j(x)$, with error small relative to the variation of the oracle across applicants.
\item[H2.] Leaf-local estimation improves on the global WoE of the block's class, and the improvement depends on the kernel exponent and the geometry.
\end{itemize}

\section{Experimental design}\label{sec:design}

\subsection{Simulated bureau data}

Evaluating an estimator of conditional evidence requires the true value of $W_j(x)$ for each applicant, which is not available in real data. The simulation is therefore constructed so that $\Prb(Y=1\mid x_S)$ can be computed for every subset of blocks. Three correlated latent factors represent financial stress $s$, credit maturity $m$ and credit seeking $q$, with $\operatorname{corr}(s,m)=-0.20$, $\operatorname{corr}(s,q)=0.35$ and $\operatorname{corr}(m,q)=-0.15$. Six characteristic blocks are generated independently given the factors (Table~\ref{tab:blocks}). The blocks are therefore dependent through the factors, and every block likelihood $p(x_k\mid s,m,q)$ has a closed form.

\begin{table}[t]
\centering\small
\caption{Characteristic blocks of the bureau-style simulation. Each block is generated from the latent factors (stress $s$, maturity $m$, credit seeking $q$) and independent noise. $\mathrm{Pois}$, $\mathrm{Bin}$ and $\mathrm{Geom}$ denote Poisson, binomial and geometric draws; $\varepsilon$ is standard normal noise drawn separately for each quantity.}
\label{tab:blocks}
\begin{tabularx}{\linewidth}{@{}p{0.17\linewidth}p{0.30\linewidth}Y@{}}
\toprule
Block & Fields & Generation given the factors\\\midrule
Revolving & limit, balance, utilization & $\log\text{limit}=\log 14000+0.45m-0.15s+0.4\varepsilon$; $\logit\text{util}=-0.7+s+0.3q+0.6\varepsilon$; balance $=$ util $\times$ limit\\
Delinquency & accounts 30+, 60+, 90+ days past due; months since delinquency & $n_{30}\sim\mathrm{Pois}(0.25e^{0.8s+0.25q})$; $n_{60}\sim\mathrm{Bin}(n_{30},0.4)$; $n_{90}\sim\mathrm{Bin}(n_{60},0.4)$; months since $\sim\min(\mathrm{Geom}(0.08),60)$ if $n_{30}>0$\\
File age & oldest and average trade age (months) & $\log(\text{oldest}-6)=\log 70+0.8m-0.15s+0.35\varepsilon$; average $=$ oldest $\times(0.25+0.55\,\mathrm{Beta}(3,2))$\\
Credit seeking & inquiries and new accounts in 6 months & inquiries $\sim\mathrm{Pois}(e^{0.1+0.4s+0.5q})$; new $\sim\mathrm{Bin}(\text{inquiries},0.35)$\\
Installment & original amount, balance, fraction outstanding & $\log\text{original}=\log18000+0.4m+0.4\varepsilon$; $\logit\text{fraction}=0.1+0.5s+0.5\varepsilon$\\
Public record & indicator & $\Prb(\text{record})=\sig(-3.3+0.8s)$\\
\bottomrule
\end{tabularx}
\end{table}

Default depends on the observed record only, through the log odds
\begin{multline}\label{eq:truth}
f^\ast(x)=-5.2+(2.6-3.4\,\tau)\,u+u^2+0.30n_{30}+0.40n_{60}+0.60n_{90}\\
+0.8\,\ind\{n_{30}>0\}e^{-\text{months since}/12}\\
+0.30\log(1+\text{inq})+0.20\,\text{new}+0.9\,\text{record}\\
+0.8\,\text{fraction}-0.25\log(\text{limit}/14000)+0.6\,\tau+u\,\ind\{n_{60}>0\},
\end{multline}
where $u$ is utilization and $\tau=\sig\{(48-\text{oldest})/8\}$ is a smooth indicator of a file younger than about four years. Holding the limit and the fields of the other blocks fixed, so that the balance changes with utilization, the derivative of the log odds with respect to utilization is
\[
\partial f^\ast/\partial u=2.6-3.4\tau+2u+\ind\{n_{60}>0\}.
\]
For established files ($\tau$ near zero) it is positive throughout. Without serious delinquency it is negative when $u<(3.4\tau-2.6)/2$, which requires $\tau>2.6/3.4$, that is, a file younger than about 39 months; as $\tau$ approaches one the condition becomes $u<0.4$. With serious delinquency the derivative is positive throughout. At 47 months, for example, $\tau=0.53$ and the derivative is $0.79+2u$, positive at every utilization. At fixed utilization, $\partial f^\ast/\partial\tau=0.6-3.4u$, so a younger file raises risk when utilization is below about 0.18 and lowers it above. These interactions are built in so that the effect of a characteristic depends on the rest of the record, which a marginal WoE cannot represent. Neither derivative equals the conditional evidence of a block, which also depends on the conditional distribution of the omitted block given the others. The through-the-door default rate is 8.7 percent, and 29 percent of records have $\tau>0.5$.

Because $\Prb(Y=1\mid x)=\sig(f^\ast(x))$, the conditional probability given a subset of blocks is
\[
\Prb(Y=1\mid x_S)=\int \sig\bigl(f^\ast(x_S,x_{-S})\bigr)\,p(x_{-S}\mid s,m,q)\,p(s,m,q\mid x_S)\,d x_{-S}\,d(s,m,q).
\]
The numerical integration, convergence checks and fitting settings are given in Appendix~\ref{app:computation}.

An earlier approval policy books an applicant with probability $s(x)=\Phi\{(1.55-\ell(x))/0.5\}$, where $\ell(x)=2u+0.5n_{30}+0.8\,\text{record}+0.3\log(1+\text{inq})$. It books 69 percent of applicants, whose default rate is 3.3 percent. It uses the revolving, delinquency, credit-seeking and public-record blocks but not file age or installment debt. An applicant is called declined by the earlier policy when $s(x)<0.5$, that is $\ell(x)>1.55$; 35 percent of test applicants fall in this group.

Five independent replicates (seeds 1 to 5) each contain 300{,}000 records divided into disjoint samples (Table~\ref{tab:samples}). Model fitting, calibration, estimation of the reference distribution and evaluation use separate records. In particular, the full model supplies projection targets on an evidence sample it did not train on; local-frequency tuning uses records outside the counts sample.

Evidence accuracy is evaluated on the union of 3{,}000 random test records and 400 sampled declines, with duplicates removed. This deliberately enriches the evaluation for the decision of interest. Its aggregate errors describe that mixture rather than the applicant population. We also report accuracy separately for the 400 sampled declines; the reference sample sets a cutoff declining the riskiest 20 percent.

\begin{table}[t]
\centering\small
\caption{Samples of each simulation replicate and their roles.}
\label{tab:samples}
\begin{tabularx}{\linewidth}{@{}lrY@{}}
\toprule
Sample & Records & Role\\\midrule
Development & 100{,}000 & production model, block-excluded models, geometry models, block classes\\
Calibration & 25{,}000 & Platt recalibration of full and outcome-fitted reduced models; choice of $(\kappa,\lambda)$ on 4{,}000 records\\
Evidence & 100{,}000 & projection regressions; leaf-local counts; $\hat\pi$\\
Reference & 50{,}000 & cutoff (riskiest 20 percent declined) and passing band (0.3 in log odds below the cutoff)\\
Test & 25{,}000 & 3{,}000 random applicants and 400 declined applicants at which evidence and attribution are evaluated\\
\bottomrule
\end{tabularx}
\end{table}

The library comparison refits the full and block-excluded models with LightGBM \citep{lightgbm} and CatBoost \citep{catboost} under the same depth, learning rate and early-stopping rule, and evaluates the same applicants and references.

\subsection{FICO HELOC data}

The FICO HELOC data provide bureau-derived characteristics and a binary outcome indicating 90 or more days past due \citep{heloc}. The OpenML version has 10{,}000 records, equally divided between good and bad outcomes; its 50 percent default rate reflects the data-set sampling rather than a portfolio. Removing 557 records without bureau information leaves 9{,}443. We exclude the consolidated bureau risk estimate and group the remaining 21 fields into delinquency (five), file age (three), trade volume and mix (four), inquiries (three), revolving burden (four) and installment burden (two).

The study imposes a score cutoff approving half of the reference sample. A HELOC ``decline'' below is a classification under this study rule, not an observed historical lending decision. Ten random splits allocate 40 percent of records to development, 10 percent to calibration, 25 percent to evidence estimation, 10 percent to references and 15 percent to test. Each declined applicant is compared with 30 references from the passing band. Applicants can recur across test splits; the split-level summaries therefore do not represent ten independent applicant samples.

Special values ($-7$, $-8$) enter the trees as ordinary values; coarse classing uses the nonnegative value and a special-value indicator for each field. This preserves the supplied codes but allows tree splits to use their ordering. Sensitivity to alternative missing-value handling is not evaluated. Without a population oracle, the HELOC analysis reports estimator agreement and local-estimation diagnostics; evidence accuracy is assessed in the simulation.

\section{Results}\label{sec:results}

\subsection{Model fit}

The production model reaches a test AUC between 0.863 and 0.869 across the five replicates, against 0.865 to 0.870 for the true log odds, and a test log loss between 0.200 and 0.209, against 0.198 to 0.208. Aggregate measures of this kind do not show how accurate the model is for an individual applicant, so the fitted log odds were compared with the true log odds directly. The root mean squared error is between 0.20 and 0.23 on the test sample and between 0.23 and 0.30 for the declined applicants. It is largest in the highest decile of true risk (0.33 to 0.40). On the hybrid records evaluated by the attribution game, which combine blocks of a declined applicant and a passing reference, the error is between 0.24 and 0.27 and its 95th percentile is between 0.47 and 0.53. These errors are small relative to the mean score gap of 1.48 but not negligible for individual blocks, which matters for the comparisons of attribution with the true model below. The passing band, from which each applicant's 50 references are drawn, contains between 2{,}199 and 2{,}564 records.

The full model and outcome-fitted reduced models use Platt recalibration; the soft-target projection regressions are not separately recalibrated. Leaf-local probabilities come from smoothed class frequencies. On the simulation test samples, the full-model calibration slopes after recalibration range from 0.97 to 1.02. This supports the probability scale used in the comparison but does not eliminate the individual log-odds errors above. Appendix~\ref{app:calibration} gives the full calibration results.

\begin{table}[t]\centering\small
\caption{Grouped baseline-Shapley attribution of the score gap for 400 declined applicants per replicate, means over five replicates. Attributions are in estimated default log odds; the mean total gap to the references is 1.48. Leading-reason shares are the fractions of applicants whose largest attribution falls on each block.}
\label{tab:aarc}
\resizebox{\linewidth}{!}{\begin{tabular}{@{}lrrrrrr@{}}\toprule
& Revolving & Delinq. & File age & Seeking & Install. & Public rec.\\\midrule
Mean attribution, production model, score band & 0.42 & 0.85 & $-$0.02 & 0.15 & 0.03 & 0.05\\
Mean attribution, true log odds, score band & 0.40 & 0.86 & $-$0.03 & 0.16 & 0.04 & 0.05\\
Mean attribution, production model, proximity & 0.46 & 0.74 & 0.03 & 0.15 & 0.03 & 0.06\\\midrule
Leading-reason share, production model & 0.26 & 0.57 & 0.05 & 0.07 & 0.00 & 0.06\\
Leading-reason share, true log odds & 0.23 & 0.56 & 0.06 & 0.08 & 0.00 & 0.07\\
Leading-reason share, proximity references & 0.33 & 0.51 & 0.01 & 0.09 & 0.01 & 0.05\\
\bottomrule\end{tabular}}
\end{table}

\subsection{Attribution against passing references}\label{sec:case}

Delinquency and revolving credit lead the score attribution for 57 and 26 percent of simulated declines, respectively. The average applicant-reference gap is 1.48 log-odds units. The fitted and true-model attributions against the same references have the same leading block for 93.2 percent of applicants (91.8--94.2 percent across replicates), and their two leading blocks overlap in 94.4 percent of positions. These comparisons isolate model fit while holding the reference records fixed (Table~\ref{tab:aarc}).

The reference policy has a larger effect than the difference between the fitted and true models. Replacing the score-band references by the most leaf-proximate references in the band changes the leading block for 32 percent of applicants, with a median Spearman correlation of 0.67 between the two attribution vectors. The most frequent changes are from delinquency to revolving credit (11 percent of applicants) and from revolving credit to delinquency (5 percent). For the applicants whose leading block changes from delinquency to revolving credit, about 45 per replicate, the proximity references have more delinquency than the score-band references (on average 1.02 against 0.61 accounts 30 or more days past due and 0.24 against 0.05 accounts 90 or more days past due) and lower utilization (0.38 against 0.53). These applicants average 1.22 delinquent accounts and utilization of 0.59. Proximity references are therefore closer to these applicants in delinquency and farther from them in utilization, which is consistent with the shift of the leading block from delinquency to revolving credit.

\begin{table}[t]\centering\small
\caption{A declined applicant from replicate 1 whose leading block changes with the reference policy. Reference columns give means over the 50 references of each policy; the share of references with any account 30+ days past due is 0.54 under the score-band policy and 1.00 under the proximity policy. The lower panel gives block attributions $C^f_j$ of the production model and the applicant's oracle evidence $W_j$ and $E_j$ (logit units).}
\label{tab:case}
\resizebox{\linewidth}{!}{\begin{tabular}{@{}lrrr@{}}\toprule
& Applicant & Score-band references & Proximity references\\\midrule
Revolving utilization & 0.53 & 0.53 & 0.27\\
Revolving limit (\$) & 30{,}809 & 15{,}936 & 29{,}493\\
Revolving balance (\$) & 16{,}229 & 8{,}176 & 8{,}089\\
Accounts 30+ days past due & 1.00 & 0.58 & 1.00\\
Accounts 60+ days past due & 1.00 & 0.20 & 0.84\\
Accounts 90+ days past due & 1.00 & 0.10 & 0.40\\
Oldest trade (months) & 294 & 81 & 181\\
Inquiries in 6 months & 1.00 & 1.98 & 1.02\\
Installment fraction outstanding & 0.36 & 0.59 & 0.43\\
Public record & 0.00 & 0.14 & 0.00\\
\midrule & Attribution, score band & Attribution, proximity & Oracle $W_j$ / $E_j$\\\midrule
Revolving & $-$0.07 & +1.14 & +1.05 / +0.71\\
Delinquency & +1.87 & +0.61 & +2.01 / +1.96\\
File age & +0.40 & 0.00 & +0.04 / $-$0.19\\
Credit seeking & $-$0.18 & +0.02 & $-$0.15 / $-$0.23\\
Installment & $-$0.13 & $-$0.01 & $-$0.17 / $-$0.56\\
Public record & $-$0.13 & 0.00 & $-$0.04 / $-$0.05\\
Total gap & 1.76 & 1.75 & \\
\bottomrule\end{tabular}}
\end{table}

The applicant in Table~\ref{tab:case} has one account 90 or more days past due, with the most recent delinquency one month earlier, and uses 53 percent of a revolving limit of \$30{,}809. Against score-band references, delinquency contributes 1.87 log-odds units to a total gap of 1.76. A contribution can exceed the gap because favorable contributions from other blocks offset it. Revolving credit contributes $-0.07$ against these references, whose mean utilization is also 53 percent and among whom 54 percent have any delinquency.

Every proximity reference has a delinquency, including 40 percent with an account 90 or more days past due. The delinquency contribution falls to 0.61. These references have mean utilization of 27 percent and a mean limit of \$29{,}493; revolving credit becomes the largest contribution at 1.14. The applicant's balance is about twice their mean balance at a similar mean limit, pointing to utilization as a candidate description. A grouped allocation does not separate utilization, balance and limit, however, and a field-level conclusion would need to preserve their defining identity.

The case was selected for the largest change in the revolving-minus-delinquency attribution among the relevant transitions, so it illustrates a pronounced rather than typical change. Its oracle conditional evidence is adverse for both revolving credit (1.05) and delinquency (2.01). Choosing which of the two passing populations is appropriate remains a reference-policy decision; the attribution accounts for the score contrast once that decision has been made.

The applicant's oldest trade is 294 months, yet file age contributes $+0.40$ against score-band references and approximately zero against proximity references. In this deliberately constructed risk model, younger files receive lower risk at high utilization. For a reference $r$, the true-model file-age allocation is $\{\tau(x)-\tau(r)\}\{0.6-1.7(u(x)+u(r))\}$, obtained by averaging whether file age enters before or after revolving credit. Averaging over the saved references gives 0.460 and 0.005 logit units, respectively, matching the exact true-model allocations. The positive fitted contribution against the score band therefore follows the constructed interaction; it is not evidence that a long credit history is generally adverse.

\begin{figure}[t]
\centering
\includegraphics[width=\linewidth]{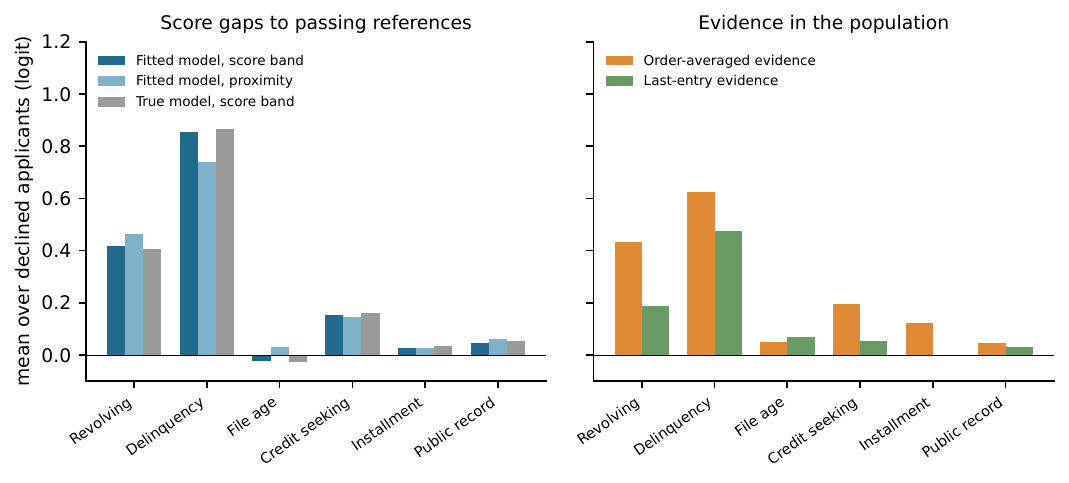}
\caption{Mean attribution of the score gap and mean oracle evidence by block for the declined applicants, averaged over five replicates. Left: score attributions against the specified passing references. Right: population evidence allocated by order averaging or measured after all other blocks are known. The panels share units but use different comparison distributions.}
\label{fig:attr}
\end{figure}

The distinction also appears in the averages. Mean revolving attribution is 0.42, close to coalition evidence of 0.43 but above last-entry evidence of 0.19. Delinquency attribution is 0.85, compared with coalition evidence of 0.62. The score-band references have more delinquency than the population average (0.62 versus 0.38 accounts 30 or more days past due), so a simple ``cleaner references'' explanation does not account for that difference; its mechanism has not been isolated. The installment attribution of 0.03 is much smaller than coalition evidence of 0.12, consistent with the counterfactual finding that most of its order-averaged evidence remains when the direct installment term is removed. Figure~\ref{fig:attr} separates the fitted-score comparisons from the population evidence allocations.

The reference-based evidence comparison of Proposition~\ref{prop:collapse} was computed with the direct estimator for every declined applicant, block and reference. The largest observed difference from the single-block swap on the production model is $7.2\times10^{-7}$, consistent with single-precision rounding of the model outputs, so the block-excluded models do not contribute to the comparison.

\subsection{Accuracy of the direct and projection estimators}

Revolving credit and delinquency provide the clearest evidence-recovery test. Their oracle evidence varies substantially across applicants, and projection recovers its sign for 99 to 100 percent of records with absolute evidence above 0.25. Across all blocks, projection has lower RMSE than separate refits: 0.09--0.18 versus 0.15--0.21 logit units (Table~\ref{tab:accuracy}). For revolving credit and delinquency, the oracle standard deviations are 0.51 and 0.78, respectively, against projection errors of 0.17 and 0.18.

Small conditional effects are harder to recover. Installment evidence has oracle standard deviation 0.10, and only 1 percent of applicants have absolute evidence above 0.25. Projection and direct estimates have correlations of 0.74 and 0.55 with the oracle. Their errors must therefore be judged against a small signal, not simply against the larger errors of marginal WoE. The zero-evidence rows of Table~\ref{tab:accuracy} make this comparison explicit.

The estimators share a full model but obtain their reduced probabilities differently: the direct method learns from binary outcomes, whereas projection learns from the full model's probabilities. Equation~\eqref{eq:error_difference} shows why the evidence error depends on both reduced-model accuracy and cancellation of shared errors. The saved predictions allow those terms to be assessed directly.

Projection improves both components of the evidence comparison. For file age, reduced-posterior RMSE falls from 0.218 to 0.190, while correlation of full and reduced errors rises from 0.625 to 0.793. For installment debt, reduced-posterior RMSE changes only from 0.211 to 0.202, but the error correlation rises from 0.738 to 0.902. The cross-product decomposition in Appendix~\ref{app:errors} confirms that greater cancellation contributes to the smaller evidence error. An accurate difference therefore does not imply equally accurate full and reduced risk estimates.

The combined evaluation sample contains 28.3 percent current-model declines, rather than the 20 percent targeted by the policy. Table~\ref{tab:decline_accuracy} reports errors separately for the 400 sampled declines per replicate. Errors are larger among declines for every block, although projection retains lower error than direct refitting.

File age illustrates what the marginal transformation misses in this constructed population. Among applicants with utilization above 50 percent, the oracle file-age evidence averages $-0.45$ for files younger than 48 months and $+0.25$ for established files. Projection gives $-0.41$ and $+0.23$, whereas global WoE averages $+0.08$ and $0.00$ for the corresponding classes. Across records with material class-level file-age evidence, global WoE has the oracle sign in only 28 percent of cases. The generating model deliberately makes younger files less adverse at high utilization: $\partial f^\ast/\partial\tau=0.6-3.4u$. Conditional evidence also depends on where the observed file age lies in its distribution given the other blocks. These group means demonstrate recovery of a designed interaction, not a claim that young credit files are generally favorable in lending.

\begin{figure}[t]
\centering
\includegraphics[width=\linewidth]{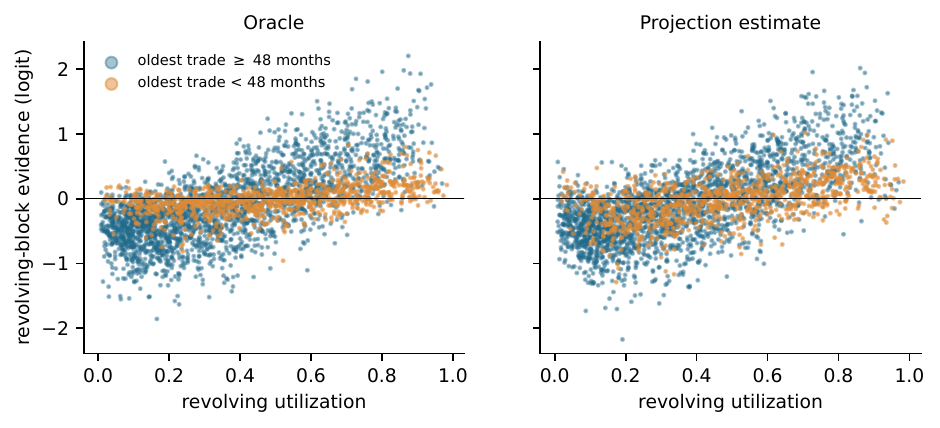}
\caption{Revolving-block conditional evidence for evaluation applicants in replicate 1. Each point is an applicant with its own remaining credit characteristics. Oracle evidence (left) varies less with utilization in the under-48-month group; projection (right) recovers the broad difference between groups with additional dispersion. Zero marks no change in default odds from observing the revolving block. These are cross-applicant patterns, not effects holding the remaining fields fixed.}
\label{fig:revolving}
\end{figure}

For revolving credit, Figure~\ref{fig:revolving} shows increasing evidence with utilization among established files and a much flatter pattern among young files. Each point is a different applicant, so its vertical position also reflects the other characteristics. The low-utilization comparison illustrates the distinction between marginal and conditional evidence: global WoE averages $-1.71$ for the class with utilization below 20 percent in both file-age groups, while oracle revolving evidence averages $-0.49$ for established files and $-0.12$ for young files. Low utilization conveys information about financial stress that delinquency and inquiries can already supply.

\begin{table}[t]\centering\small
\caption{Error of conditional evidence estimates against the oracle in the bureau-style simulation: root mean squared error (logit units), mean over five replicates, for about 3{,}350 combined evaluation applicants per replicate. Direct and projection estimates are compared with the pointwise evidence $W_j$ and the global and leaf-local estimates with the class-level evidence $W^B_j$; errors in different panels refer to different targets and are not compared with each other. The first two rows give the standard deviation of the oracle across applicants. Zero-evidence baselines assign zero to every applicant for each target. Replicate standard deviations of the RMSE are at most 0.02 for direct and projection estimates and at most 0.07 for leaf-local estimates.}
\label{tab:accuracy}
\resizebox{\linewidth}{!}{\begin{tabular}{@{}lrrrrrr@{}}\toprule
& Revolving & Delinq. & File age & Seeking & Install. & Public rec.\\\midrule
Oracle sd, pointwise $W_j$ & 0.51 & 0.78 & 0.31 & 0.29 & 0.10 & 0.20\\
Oracle sd, class-level $W^B_j$ & 0.46 & 0.68 & 0.29 & 0.26 & 0.09 & 0.20\\\midrule
Zero evidence, pointwise target & 0.51 & 0.79 & 0.31 & 0.29 & 0.10 & 0.20\\
Direct & 0.20 & 0.21 & 0.19 & 0.17 & 0.15 & 0.16\\
Projection & 0.17 & 0.18 & 0.13 & 0.13 & 0.09 & 0.10\\\midrule
Zero evidence, class-level target & 0.47 & 0.69 & 0.29 & 0.26 & 0.09 & 0.20\\
Global WoE of class & 0.97 & 0.59 & 0.38 & 0.63 & 0.78 & 0.12\\
Leaf-local, default geometry, $\kappa=1$ & 0.72 & 0.42 & 0.33 & 0.46 & 0.61 & 0.10\\
Leaf-local, default geometry, selected $\kappa$ & 0.27 & 0.16 & 0.28 & 0.19 & 0.23 & 0.07\\
Leaf-local, class geometry, $\kappa=1$ & 0.71 & 0.42 & 0.33 & 0.40 & 0.53 & 0.08\\
Leaf-local, class geometry, selected $\kappa$ & 0.24 & 0.16 & 0.22 & 0.14 & 0.18 & 0.05\\
\midrule
Sign agreement$^{a}$, projection & 0.99 & 1.00 & 0.99 & 0.99 & 0.96 & 1.00\\
Sign agreement, global WoE & 0.90 & 0.96 & 0.28 & 0.97 & -- & 1.00\\
Sign agreement, leaf-local (class, selected) & 0.99 & 1.00 & 0.94 & 0.99 & -- & 1.00\\
Share with $|W_j|>0.25$ & 0.55 & 0.76 & 0.24 & 0.37 & 0.01 & 0.05\\
\bottomrule\end{tabular}}
\par\smallskip\raggedright\footnotesize $^{a}$Share of applicants with $|$oracle$|>0.25$ whose estimate has the oracle's sign; -- indicates 20 or fewer such applicants. The qualifying records can differ between the two targets.
\end{table}

\subsection{Accuracy of the leaf-local estimator}

The leaf-local estimator targets the class-level evidence $W^B_j$, which differs from the pointwise evidence by an RMSE of 0.21 for revolving credit and 0.40 for delinquency because the block classes are coarse. Its errors are therefore compared with those of global WoE, the marginal class-level WoE used as a baseline and evaluated against the same conditional class-level oracle, and not with the pointwise estimators. With the proximity kernel of Eq.~\eqref{eq:kernel} used as originally specified ($\kappa=1$), the estimator improves on global WoE for every block, but its error remains larger than the oracle standard deviation for the revolving, file-age, credit-seeking and installment blocks (Table~\ref{tab:accuracy}). The effective sample sizes indicate how broadly the records are weighted. With $\kappa=1$, the median default-class effective sample size is between 5{,}992 and 7{,}133 in an evidence sample containing about 8{,}700 defaults, so the weights spread over most of the sample and the estimate stays close to the global value. Each root-to-leaf path of a depth-4 tree has at most four splits, and averaging leaf membership over trees does not restrict all the other fields at once.

\begin{figure}[t]
\centering
\includegraphics[width=\linewidth]{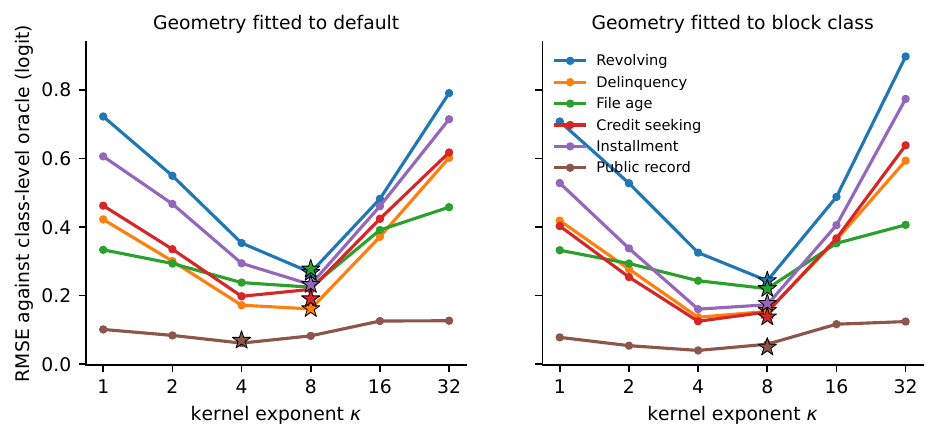}
\caption{Leaf-local RMSE against the class-level oracle across kernel exponents. Lines minimize oracle error over the smoothing constant after averaging across replicates; they describe the error surface rather than a deployable selection rule. Stars show mean RMSE at the held-out-likelihood-selected exponent and smoothing constant, placed at the modal selected exponent for each block. $\kappa=1$ is the unsharpened proximity average.}
\label{fig:kernel}
\end{figure}

Raising the kernel to a power concentrates the weights (Figure~\ref{fig:kernel}). The held-out likelihood criterion selected $\kappa=8$ in 22 of the 30 block-replicate combinations for the default geometry, $\kappa=4$ in 6 and $\kappa=16$ in 2, and selected the smallest smoothing constant, $\lambda=0.1$, in 27 of 30. The oracle error is lowest at $\kappa=8$ for four blocks and at $\kappa=4$ for the other two, so the criterion, which uses no oracle information, selects at or next to the best exponent. At the selected exponent, the median default-class effective sample size falls to between 110 and 814 and the error falls by a factor of 1.2 (file age) to 2.7 (revolving credit) relative to $\kappa=1$. Larger exponents increase the error again as the effective sample size of defaulters becomes small. Within a block, the error decreases with the local effective sample size: for revolving credit, the RMSE is 0.32 in the lowest tercile of default-class effective sample size and 0.23 in the highest.

At the selected exponent, class-prediction neighborhoods have lower error than default-prediction neighborhoods for every block, most noticeably file age (0.22 versus 0.28) and installment debt (0.18 versus 0.23). The class geometry better recovers the held-out class frequencies in this experiment. Yet the installment error of 0.18 remains larger than the 0.095 RMSE obtained by assigning zero class-level evidence to every applicant. H2 is supported as an improvement over marginal WoE, particularly with concentrated weights; that improvement does not establish accurate recovery of every block's conditional evidence.

Table~\ref{tab:samplesize} repeats the class-geometry estimate for replicate 1 with smaller evidence samples drawn from the same population. The error rises as the sample shrinks, from 0.22 to 0.62 for revolving credit and from 0.16 to 0.63 for installment between 100{,}000 and 2{,}500 records. Even with 2{,}500 records the estimate has lower error than the global WoE for every block. The experiment varies only the evidence sample; the geometry models and the calibration records are unchanged.

\begin{table}[t]\centering\small
\caption{Leaf-local estimation with evidence samples of different sizes, replicate 1. RMSE against the class-level oracle for the class geometry at the selected $(\kappa,\lambda)$, and for global WoE computed from the same subsample. The last row gives the median default-class effective sample size at 100{,}000 records.}
\label{tab:samplesize}
\resizebox{\linewidth}{!}{\begin{tabular}{@{}lrrrrrr@{}}\toprule
Evidence records & Revolving & Delinq. & File age & Seeking & Install. & Public rec.\\\midrule
2{,}500 & 0.62 & 0.32 & 0.41 & 0.36 & 0.63 & 0.06\\
10{,}000 & 0.51 & 0.19 & 0.39 & 0.22 & 0.32 & 0.07\\
30{,}000 & 0.36 & 0.17 & 0.29 & 0.16 & 0.20 & 0.11\\
100{,}000 & 0.22 & 0.14 & 0.23 & 0.15 & 0.16 & 0.08\\
Global WoE, 100{,}000 & 0.92 & 0.58 & 0.39 & 0.62 & 0.77 & 0.13\\
Global WoE, 2{,}500 & 1.01 & 0.66 & 0.45 & 0.71 & 0.98 & 0.14\\
Default-class ESS, 100{,}000 & 170 & 171 & 357 & 271 & 1061 & 271\\
\bottomrule\end{tabular}}
\end{table}

\subsection{Additivity and coalition evidence}

The leave-one-block-out values leave much of the total evidence unallocated for most declined applicants. Their mean sum is 0.74--0.88 across replicates, compared with total evidence of 1.39--1.58. The sum falls below the total by more than 0.05 for 75--81 percent of applicants and exceeds it by more than 0.05 for 15--21 percent. These are signs of an allocation discrepancy, not counts of redundant and synergistic applicants. Between 1 and 3 percent of model declines have negative population evidence, so decline status alone does not imply risk above the population average.

Order averaging allocates the total and yields projection errors of 0.05--0.18 for $E_j$, against oracle standard deviations of 0.28--1.14 (Table~\ref{tab:coalition}). The mean excess over last-entry evidence is largest for revolving credit (0.24), delinquency (0.15), credit seeking (0.14) and installment debt (0.12), and slightly negative for file age. The installment counterfactual allows a more specific interpretation than these signed averages alone.

The installment block illustrates the difference between applicant-level and order-averaged evidence. Over the test applicants, its leave-one-block-out evidence has standard deviation 0.10 and mean absolute value 0.08, and over the declined applicants its mean is 0.00 with standard deviation 0.09: the direct term $0.8\times\text{fraction}$ in Eq.~\eqref{eq:truth} produces evidence of both signs that averages out. To separate the direct term from the association of the installment fraction with financial stress, the oracle was recomputed for 1{,}000 test applicants and 200 declined applicants of replicate 1 with the direct term removed. The leave-one-block-out evidence of the installment block is then exactly zero, while its mean coalition evidence for the declined applicants falls only from 0.129 to 0.116. Most of the block's average coalition evidence for these applicants therefore comes from its association with the other blocks, and its applicant-level conditional evidence comes from the direct term.

\begin{table}[t]\centering\small
\caption{Coalition evidence of the 400 declined applicants per replicate and the effect of the earlier approval policy, means over five replicates (logit units). The coalition excess is $H_j=E_j-W_j$. The absolute booked-minus-through-the-door difference $|W^b_j-W_j|$ is computed from the oracle for the test applicants; earlier-policy declines are applicants with booking probability $s(x)<0.5$ and current-model declines are applicants with $f(x)>c$. The last three rows give the RMSE of direct estimates fitted on booked accounts only, against the through-the-door and booked oracles, and of direct estimates fitted on all applicants against the through-the-door oracle.}
\label{tab:coalition}
\resizebox{\linewidth}{!}{\begin{tabular}{@{}lrrrrrr@{}}\toprule
& Revolving & Delinq. & File age & Seeking & Install. & Public rec.\\\midrule
Coalition evidence $E_j$, oracle mean & 0.43 & 0.62 & 0.05 & 0.19 & 0.12 & 0.05\\
Leave-one-block-out $W_j$, oracle mean & 0.19 & 0.47 & 0.07 & 0.05 & 0.00 & 0.03\\
Coalition excess $H_j$, oracle mean & 0.24 & 0.15 & $-$0.02 & 0.14 & 0.12 & 0.01\\
RMSE of projection estimate of $E_j$ & 0.09 & 0.18 & 0.07 & 0.08 & 0.05 & 0.06\\
Oracle sd of $E_j$ & 0.68 & 1.14 & 0.28 & 0.57 & 0.35 & 0.35\\\midrule
$|W^b_j-W_j|$, all test applicants & 0.23 & 0.35 & 0.00 & 0.08 & 0.00 & 0.05\\
$|W^b_j-W_j|$, earlier-policy declines & 0.47 & 0.75 & 0.00 & 0.21 & 0.00 & 0.09\\
$|W^b_j-W_j|$, current-model declines & 0.54 & 0.74 & 0.00 & 0.22 & 0.00 & 0.09\\
Direct, booked fit vs.\ through-the-door oracle & 0.40 & 0.50 & 0.27 & 0.26 & 0.22 & 0.25\\
Direct, booked fit vs.\ booked oracle & 0.31 & 0.40 & 0.27 & 0.26 & 0.22 & 0.26\\
Direct, all-applicant fit vs.\ through-the-door oracle & 0.20 & 0.21 & 0.19 & 0.17 & 0.15 & 0.16\\
\bottomrule\end{tabular}}
\end{table}

\subsection{Booked-population evidence}

Booking changes the comparison distribution for revolving credit, delinquency, credit seeking and public records, the four blocks used by the earlier policy. Their mean booked-minus-through-the-door evidence differences are positive. For revolving credit and delinquency they average 0.23 and 0.35 over the evaluation sample, rising to 0.54 and 0.74 among current-model declines (Table~\ref{tab:coalition}). Among applicants with earlier-policy booking probability below 0.5, the corresponding means are 0.47 and 0.75. The two decline groups overlap substantially: 93 percent of current-model declines also have booking probability below 0.5.

File age and installment debt are absent from that policy, so their population evidence differences are zero up to numerical error. The sign for the policy's blocks is an empirical average, not a universal consequence of the booking rule: 11 revolving estimates across five replicates are slightly negative, with minimum $-0.0008$. The conditional covariance condition of Proposition~\ref{prop:selection} was not verified, and the model's utilization effect is not monotone for the youngest files.

Direct estimates fitted on booked accounts are closer to the booked oracle than to the through-the-door oracle: revolving RMSE is 0.31 versus 0.40, and delinquency RMSE is 0.40 versus 0.50. They also have larger error than all-applicant fits for blocks unaffected by population selection. The smaller booked sample and its 3.3 percent default rate accompany that loss of accuracy. The main selection finding is that booked-population evidence represents a different conditional comparison, especially for applicants classified as declines.

\subsection{Conditional information value}\label{sec:civ_results}

Table~\ref{tab:civ} reports conditional information values. In the simulation the joint information value of the record is 2.51. The marginal values sum to 4.56, nearly twice that, because revolving credit, delinquency, credit seeking and installment debt all carry information about the same stress factor. The leave-one-block-out information values sum to 1.43, and the coalition values add up exactly to the joint value. Installment debt shows the difference most clearly: its marginal value is 0.68, but once the other blocks are known it adds 0.01, and its coalition value of 0.24 is its share of the information it holds in common with them. File age moves the other way, from 0.07 marginally to 0.16 conditionally, because its effect reverses with utilization and largely cancels in the population. Projection estimates recover the oracle values within 0.03 for both $\mathrm{IV}_j$ and $\mathrm{IV}^E_j$, and the gap in mean fitted log odds between defaulters and non-defaulters gives 2.45 for the joint value.

In HELOC the estimated joint value is 1.32, against 1.83 for the sum of marginal values and 1.02 for the sum of leave-one-block-out values. Revolving burden and delinquency carry the most information given the rest of the record. Trade volume falls from 0.24 marginally to 0.04 conditionally, so most of what it shows is already in the other blocks. Because conditional evidence is a log likelihood ratio, these values are not affected by the 50 percent default rate of the sample.

\begin{table}[t]\centering\small
\caption{Conditional information value (natural units). Simulation: oracle and projection values on 1{,}000 random test applicants per replicate, class means weighted by the true default probability; means over 5 replicates, replicate standard deviations at most 0.10 for the oracle values. HELOC: projection estimates over the test records, means over 10 splits. Marginal values use the evidence of the block alone, $V_x(\{j\})$, or its projection estimate.}
\label{tab:civ}
\resizebox{\linewidth}{!}{\begin{tabular}{@{}lrrrrrr@{}}\toprule
\multicolumn{7}{@{}l}{\emph{Simulation}}\\
 & Revolving & Delinq. & File age & Seeking & Install. & Public rec.\\\midrule
Marginal IV, oracle & 1.43 & 1.51 & 0.07 & 0.75 & 0.68 & 0.12\\
Conditional IV $\mathrm{IV}_j$, oracle & 0.32 & 0.80 & 0.16 & 0.08 & 0.01 & 0.05\\
Coalition IV $\mathrm{IV}^E_j$, oracle & 0.74 & 1.04 & 0.11 & 0.32 & 0.24 & 0.07\\
Conditional IV $\mathrm{IV}_j$, projection & 0.31 & 0.79 & 0.14 & 0.08 & 0.01 & 0.04\\
Coalition IV $\mathrm{IV}^E_j$, projection & 0.72 & 1.04 & 0.10 & 0.30 & 0.23 & 0.06\\
\midrule
Joint IV $\E[f^\ast(X)\mid Y=1]-\E[f^\ast(X)\mid Y=0]$ & \multicolumn{6}{c}{2.51 (projection 2.45)}\\
Sum of marginal / conditional / coalition IVs, oracle & \multicolumn{6}{c}{4.56 / 1.43 / 2.51}\\
\midrule
\multicolumn{7}{@{}l}{\emph{FICO HELOC}}\\
 & Delinquency & File age & Trade vol. & Inquiries & Revolving & Installment\\\midrule
Marginal IV, projection & 0.40 & 0.26 & 0.24 & 0.28 & 0.61 & 0.04\\
Conditional IV $\mathrm{IV}_j$, projection & 0.31 & 0.13 & 0.04 & 0.15 & 0.36 & 0.02\\
Coalition IV $\mathrm{IV}^E_j$, projection & 0.35 & 0.18 & 0.12 & 0.19 & 0.46 & 0.02\\
\midrule
Joint IV, projection & \multicolumn{6}{c}{1.32}\\
Sum of marginal / conditional / coalition IVs & \multicolumn{6}{c}{1.83 / 1.02 / 1.32}\\
\bottomrule\end{tabular}}
\end{table}
\FloatBarrier

\subsection{FICO HELOC data}

The HELOC model has test AUC 0.782--0.807 across splits. Its fitted Platt slope ranges from 0.81 to 1.18; test calibration slopes range from 0.87 to 1.14 and intercepts from $-0.19$ to $0.08$. The raw margins have slopes of 0.91--1.08, and one recalibration moves a test slope from 0.91 to 0.87. These ranges describe small-sample calibration uncertainty rather than evidence that recalibration uniformly improves the predictions. The calibration sample contains about 940 records, and the evidence sample about 2{,}360. Passing bands contain 45--76 records, from which each applicant's 30 references are drawn.

Direct and projection evidence are correlated most strongly for delinquency (0.84) and revolving burden (0.87), with inquiry correlation 0.76. Agreement is weaker for trade volume (0.50) and installment burden (0.35), the two blocks with the least estimated evidence (Table~\ref{tab:heloc}). These methods share the full model, so their agreement is not an independent accuracy check.

The leaf-local estimates correlate with global WoE at 0.69 to 0.95 for five of six blocks and with the projection estimates at 0.04 to 0.76. The class-balanced criterion selected $\kappa=4$ in 44 of the 60 block-split combinations, $\kappa=8$ in 11 and $\kappa=2$ in 5, and $\lambda=0.1$ in 54. At these settings the median default-class kernel-weighted count is between 7 and 26 and the median default-class effective sample size between 88 and 251. The mean absolute difference between the leaf-local estimate and the global WoE of the class is between 0.13 and 0.28, against standard deviations of the global WoE across applicants of 0.18 to 0.81. The leaf-local estimates therefore depart from global WoE, but without an oracle it cannot be determined whether the departures are accurate. The low correlation with the projection estimates for trade volume and installment burden is more consistent with sampling noise dominating a weak signal than with a failure of the leaf geometry: these two blocks carry the least information given the rest of the record (conditional information values 0.04 and 0.02, Table~\ref{tab:civ}), although the data cannot separate the two. The simulation shows that 2{,}500 evidence records can be enough for the leaf-local estimator to improve on global WoE in that population. It does not establish a sample-size requirement for HELOC, and the HELOC results provide limited evidence about leaf-local estimation in the configuration studied.

The estimated leave-one-block-out evidence sums to less than the estimated total evidence for 45 to 61 percent of declined applicants and to more for 36 to 50 percent. The estimated coalition excess is small for every block. For delinquency and revolving burden, which together lead for 71 percent of declined applicants, the mean attribution (0.37 and 0.36) is close to both the coalition evidence (0.30 and 0.37) and the leave-one-block-out evidence (0.30 and 0.33).

\begin{table}[t]\centering\small
\caption{FICO HELOC data, ten random splits. Correlations are across test applicants within a split, averaged over splits. Means of $E_j$, $W_j$ and the attribution $C^f_j$ are over declined applicants. Estimated evidence uses the projection estimator.}
\label{tab:heloc}
\resizebox{\linewidth}{!}{\begin{tabular}{@{}lrrrrrr@{}}\toprule
& Delinq. & File age & Trades & Inquiries & Revolving & Install.\\\midrule
Corr(direct, projection) & 0.84 & 0.72 & 0.50 & 0.76 & 0.87 & 0.35\\
Corr(projection, leaf-local default geometry) & 0.75 & 0.58 & 0.14 & 0.48 & 0.76 & 0.04\\
Corr(leaf-local default geometry, global WoE) & 0.93 & 0.89 & 0.69 & 0.92 & 0.95 & 0.46\\\midrule
Mean coalition evidence $E_j$ & 0.30 & 0.15 & 0.09 & 0.13 & 0.37 & 0.02\\
Mean leave-one-block-out $W_j$ & 0.30 & 0.14 & 0.04 & 0.13 & 0.33 & 0.04\\
Mean attribution $C^f_j$ & 0.37 & 0.22 & 0.11 & 0.18 & 0.36 & 0.01\\
Leading-reason share & 0.39 & 0.20 & 0.04 & 0.06 & 0.32 & 0.00\\
\bottomrule\end{tabular}}
\end{table}

\section{Discussion}\label{sec:discussion}

Conditional evidence gives a credit analyst a comparison that marginal WoE cannot supply: how the observed characteristic changes default odds after the rest of this record is known. The numerical example's increase from 10 to 20 percent risk corresponds to 0.81 log-odds units. In the simulation, the comparison recovers a deliberately constructed file-age interaction that a single marginal class value misses. It also shows why a small signed average cannot establish that a block is irrelevant. Installment evidence varies across applicants, and removing its direct risk term eliminates its last-entry evidence while leaving most of its order-averaged evidence.

The estimator must be judged against the conditional evidence for each applicant. Averaging estimation errors across applicants assesses how accurately these local quantities are recovered; it does not turn them into a measure of overall model fit. Projection has lower error than separate refits in this simulation, with both smaller reduced-posterior errors and more cancellation of errors shared with the full model. This makes it useful for describing the evidence implied by an accurate model, but does not establish robustness to a misspecified full model. Leaf-local frequencies provide a different, coarser comparison. Concentrating their weights improves on marginal WoE, yet the installment estimate remains less accurate than assigning zero conditional evidence. The amount of recoverable evidence therefore matters as much as the improvement over a weak baseline.

For adverse action, the relevant scored comparison remains the declined applicant against specified passing records. In the featured case, delinquency leads against the broader score band and revolving credit leads against references with more similar delinquency histories. Both comparisons reconcile the score gap. They identify different disadvantages because the reference populations differ. A lender interpreting the leading block must therefore consider the reference characteristics and whether the block description accurately represents the scored factors. Exact allocation does not choose the reference policy or resolve field-level wording within a multi-field block.

The population being described is also consequential. Booked accounts can have a different distribution of an omitted characteristic from through-the-door applicants with the same remaining record. In the selection experiment, booked-population evidence is larger on average for the blocks used by the earlier policy, with revolving and delinquency differences of 0.54 and 0.74 logit units among current-model declines. Those numbers describe the studied policy and population; selection on information outside the observed record would also alter the full-record posterior.

The simulation was designed to expose conditional effects and permits numerical assessment of evidence error. It does not reproduce all bureau missingness, monthly histories or population changes. Its full model is well fitted, so the projection advantage has not been tested under systematic model misspecification or in a generating model without the constructed reversal. The oracle sensitivity checks produce small discrepancies between numerical approximations, but cannot detect bias common to those approximations. Six blocks permit exact order averaging; approximation error with many more blocks is not studied.

HELOC supplies a real-data application under an imposed approval rule, without a population oracle or observed notice reasons. Its calibration, special-value handling and small local samples can affect evidence estimates, but the separate contribution of each was not identified. The attribution hybrids preserve within-block identities but were not checked for cross-block support. Regulatory interpretation is confined to the cited Regulation B text and Official Interpretations; these do not specifically address grouped Shapley attribution.

\section{Conclusion}

Conditional weight of evidence measures the adverse or favorable information a characteristic adds for an individual applicant, given the remaining credit record. In this simulation, regression of full-model probabilities estimates that information more accurately than separate outcome-based refits, given an accurately specified full model. The kernel exponent of the leaf-local estimator chosen by held-out likelihood, without oracle information, is at or next to the exponent with the lowest oracle error. Order averaging provides a related allocation of total evidence across characteristics, including information they share. Averaged within each outcome class, the same evidence gives a conditional information value: the expected conditional evidence for default among defaulters plus that against default among non-defaulters, equal to the gap between their average conditional WoE. It is on the WoE scale, it is never negative, and its order-averaged version divides the information value of the whole model among the characteristics.

Passing-reference choice changes the leading attributed block for about a third of simulated declines. Interpreting a characteristic's local risk information and explaining its contribution to a score gap therefore involve distinct comparisons. Conditional WoE supplies the first; the second requires a specified passing population and an allocation of the fitted score difference.

\clearpage
\appendix
\section{Numerical integration and experimental settings}\label{app:computation}

The population evidence requires integration over unobserved characteristics. The following checks assess sensitivity to the numerical approximation, while the fitting settings specify the models whose evidence and attributions are compared.

The latent posterior $p(s,m,q\mid x_S)$ is evaluated on a grid of $34^3$ points spanning $\pm4.3$ standard deviations, and the integral is computed by drawing 8{,}192 latent values from the grid posterior (with uniform jitter inside each cell) and one draw of the absent blocks for each. Two checks support the oracle. The probability integral transform of the true latent values under the grid posterior has mean between 0.47 and 0.51 and standard deviation between 0.28 and 0.30 for each factor, close to the uniform values of 0.5 and 0.289. A logistic regression of fresh outcomes on the oracle $\logit\Prb(Y=1\mid x_{-j})$ has slope between 0.98 and 1.01 for every block. The same draws give the class-level oracle $W^B_j$ of Eq.~\eqref{eq:binwoe}, computed from the Monte Carlo frequencies of the applicant's class among the draws, and the booked-population oracle of Eq.~\eqref{eq:booked}.

These checks do not bound the error for an individual applicant on the logit scale, so parts of the oracle were recomputed for replicate 1. The leave-one-block-out and class-level evidence were recomputed for 300 random test applicants outside the declined sample, using both independent draws at the original settings and 32{,}768 draws on a $48^3$ grid spanning $\pm5$ standard deviations. Coalition evidence was recomputed for 100 declined applicants using the finer settings. With the finer computation, the root mean squared change in $W_j$ is between 0.001 and 0.011 logit units across blocks and the 99th percentile of the absolute change is at most 0.030; for $W^B_j$ the corresponding values are at most 0.011 and 0.029, and for the coalition evidence $E_j$ at most 0.005 and 0.014. The independent replicate at the original settings changes $W_j$ by at most 0.015 in root mean square. These changes are small relative to the estimator errors in the main results, which are at least 0.05. The smallest Monte Carlo probability of an applicant's class among these applicants is $4.9\times10^{-4}$, about four draws, and no class probability was zero. Across all saved oracle values, the conditional default probabilities lie between 0.004 and 0.9995 for the leave-one-block-out calculation and between 0.009 and 0.9995 for the coalition calculation, so clipping at $10^{-4}$ and $1-10^{-4}$ is not active for them. These are discrepancies between numerical calculations on the checked subsets, not bounds on the error relative to the exact integrals, and an approximation bias shared by both calculations would not be detected. The oracle values are numerical estimates of the population quantities.

The primary risk, projection and neighborhood models use XGBoost \citep{xgboost} using trees of depth 4, learning rate 0.05, row subsampling 0.8 and early stopping on a random 20 percent of its training sample. The multiclass geometry models use learning rate 0.1. The kernel exponent is chosen from $\{1,2,4,8,16,32\}$ and the smoothing constant from $\{0.1,0.5,2,8,32\}$. Each block has five classes except delinquency, which has three because most records have no delinquent account, and public record, which has two. Each declined applicant is compared with 50 references. A separate run for replicate 1 recomputes the leaf-local estimates with evidence samples of 2{,}500, 10{,}000, 30{,}000 and 100{,}000 records drawn from the same evidence sample, with the geometry models and calibration records unchanged.
\section{Calibration and library comparisons}\label{app:calibration}

Calibration describes the probability scale on which the evidence estimates depend. The full and direct reduced models are Platt-recalibrated on a separate sample; projection models learn those full-model probabilities without separate outcome recalibration. The production model's test diagnostics are reported in Table~\ref{tab:calibration}.

In the simulation, the unrecalibrated margin has test calibration slope 1.00--1.02 and the fitted Platt slope is 1.00--1.03. After recalibration the test slope is 0.97--1.02, the intercept is $-0.03$ to $0.02$ and decile expected calibration error is at most 0.005. Mean bias against true log odds ranges from $-0.025$ to $0.047$, with regression slope of true on fitted log odds 0.99--1.01. Among declines, predicted default probability averages 0.32--0.33 versus true probability 0.32, and the true-on-fitted log-odds slope is 1.04--1.08. These aggregate agreements coexist with the applicant-level errors in the main text.

\begin{table}[t]\centering\small
\caption{Calibration of the production model on the test sample: ranges over five simulation replicates and ten HELOC splits. The calibration slope and intercept are from a logistic regression of the outcome on the model's log odds (intercept with slope fixed at one). The lower panel compares the simulation model with the true log odds, for all test applicants and for applicants above the cutoff.}
\label{tab:calibration}
\begin{tabular}{@{}lrr@{}}\toprule
& Simulation & HELOC\\\midrule
Platt slope $b$ & 1.00 to 1.03 & 0.81 to 1.18\\
Calibration slope before Platt & 1.00 to 1.02 & 0.91 to 1.08\\
Calibration slope & 0.97 to 1.02 & 0.87 to 1.14\\
Calibration intercept & $-$0.03 to 0.02 & $-$0.19 to 0.08\\
Expected calibration error (deciles) & 0.002 to 0.005 & 0.021 to 0.047\\
Mean predicted probability & 0.087 to 0.092 & 0.484 to 0.528\\
Observed default rate & 0.086 to 0.091 & 0.478 to 0.512\\
\midrule
Bias against true log odds, all & $-$0.025 to 0.047 & --\\
Slope of true on fitted log odds, all & 0.991 to 1.012 & --\\
Bias against true log odds, declined & $-$0.008 to 0.049 & --\\
Slope of true on fitted log odds, declined & 1.039 to 1.076 & --\\
Mean predicted probability, declined & 0.321 to 0.332 & --\\
Mean true probability, declined & 0.317 to 0.323 & --\\
\bottomrule\end{tabular}
\end{table}

\subsection{Gradient-boosting libraries}

LightGBM and CatBoost fitted with the same depth, learning rate and stopping rule reach test AUCs of 0.862 to 0.869 and 0.863 to 0.869. Their direct conditional evidence estimates have RMSE between 0.17 and 0.21 (LightGBM) and between 0.12 and 0.16 (CatBoost) across blocks, comparable to the 0.15 to 0.21 of XGBoost. Against the same applicants and references, the leading candidate reason agrees with the XGBoost model's leading reason for 95 to 97 percent of applicants under LightGBM and 94 to 96 percent under CatBoost, and with the true model's leading reason for 92 to 96 and 93 to 95 percent. Of the XGBoost score-band references, 87 percent pass under the LightGBM model and 78 to 82 percent under the CatBoost model at the same approval rate, so the choice of library changes which records qualify as passing references more than it changes the attribution against a fixed reference set.

\section{Zero-evidence baselines and error decomposition}\label{app:errors}

The zero-evidence baseline has RMSE $\sqrt{\E[W_j^2]}$ for the pointwise target and $\sqrt{\E[(W_j^B)^2]}$ for the class-level target. It measures the error of assigning no conditional evidence to any applicant. Table~\ref{tab:error_decomposition} evaluates Eq.~\eqref{eq:error_difference} on the same combined evaluation records as the main accuracy table. Entries are means of replicate mean squared errors and cross-products, so the identity holds across each row before rounding. Positive cross-products represent cancellation in the evidence difference; correlation alone does not quantify that contribution.
\begin{table}[t]\centering\small
\caption{Error in the evidence difference, averaged over five replicates on the combined evaluation sample. The first two terms are squared full and reduced log-odds errors; subtracting twice their cross-product gives the evidence MSE. Values are in squared log-odds units.}
\label{tab:error_decomposition}
\begin{tabular}{@{}llrrrr@{}}\toprule
Block & Estimator & $\E[e_F^2]$ & $\E[e_{-j}^2]$ & $2\E[e_Fe_{-j}]$ & Evidence MSE\\\midrule
Revolving & Direct & 0.0477 & 0.0405 & 0.0490 & 0.0392\\
 & Projection & 0.0477 & 0.0297 & 0.0490 & 0.0284\\
Delinquency & Direct & 0.0477 & 0.0375 & 0.0412 & 0.0440\\
 & Projection & 0.0477 & 0.0303 & 0.0439 & 0.0341\\
File age & Direct & 0.0477 & 0.0476 & 0.0602 & 0.0351\\
 & Projection & 0.0477 & 0.0361 & 0.0662 & 0.0177\\
Credit seeking & Direct & 0.0477 & 0.0424 & 0.0628 & 0.0273\\
 & Projection & 0.0477 & 0.0360 & 0.0672 & 0.0165\\
Installment & Direct & 0.0477 & 0.0446 & 0.0686 & 0.0236\\
 & Projection & 0.0477 & 0.0411 & 0.0800 & 0.0088\\
Public record & Direct & 0.0477 & 0.0428 & 0.0650 & 0.0255\\
 & Projection & 0.0477 & 0.0401 & 0.0770 & 0.0108\\
\bottomrule\end{tabular}
\end{table}
\begin{table}[t]\centering\small
\caption{Pointwise conditional-evidence RMSE among the 400 sampled declines per replicate, compared with the combined evaluation sample. Entries are means of replicate RMSE in logit units. The decline sample is selected by the original production model and cutoff.}
\label{tab:decline_accuracy}
\resizebox{\linewidth}{!}{\begin{tabular}{@{}lrrrrrr@{}}\toprule
& Revolving & Delinq. & File age & Seeking & Install. & Public rec.\\\midrule
Direct, combined & 0.198 & 0.210 & 0.187 & 0.165 & 0.153 & 0.159\\
Direct, sampled declines & 0.236 & 0.278 & 0.232 & 0.195 & 0.190 & 0.195\\
Projection, combined & 0.168 & 0.185 & 0.133 & 0.128 & 0.093 & 0.104\\
Projection, sampled declines & 0.190 & 0.263 & 0.169 & 0.158 & 0.108 & 0.143\\
\bottomrule\end{tabular}}
\end{table}

\end{document}